\documentclass[11pt,a4paper]{article}
\usepackage[hyperref]{acl2021}
\usepackage{times}
\usepackage{latexsym}

\usepackage{microtype}

\usepackage{url}

\usepackage{amsmath}
\usepackage{amssymb}
\usepackage{graphicx}
\usepackage{multirow}
\usepackage{booktabs}

\newcommand{\R}{\mathbb{R}}

\makeatletter
\newcommand{\ssymbol}[1]{$^{\@fnsymbol{#1}}$}
\makeatother
\let\bm

\let\boldsymbol

\newlength\savewidth

\aclfinalcopy 

\title{O2NA: An Object-Oriented Non-Autoregressive Approach for \\ Controllable Video Captioning}

\author{Fenglin Liu\textsuperscript{1}\thanks{\ \ Equal Contributions.}, Xuancheng Ren\textsuperscript{2}\footnotemark[1], Xian Wu\textsuperscript{4}, Bang Yang\textsuperscript{1}, Shen Ge\textsuperscript{4}, Yuexian Zou\textsuperscript{1,5}, Xu Sun\textsuperscript{2,3}\\
\textsuperscript{1}ADSPLAB, School of ECE, Peking University\\
\textsuperscript{2}MOE Key Laboratory of Computational Linguistics, School of EECS, Peking University\\
\textsuperscript{3}Center for Data Science, Peking University\\
\textsuperscript{4}Tencent, Beijing, China \ \  \textsuperscript{5}Peng Cheng Laboratory, Shenzhen, China\\
{\tt \{fenglinliu98, renxc, yb.ece, zouyx, xusun\}@pku.edu.cn}\\  {\tt \{kevinxwu, shenge\}@tencent.com}\\ 
}

\date{}

\begin{document}
\maketitle
\begin{abstract}
Video captioning combines video understanding and language generation. Different from image captioning that describes a static image with details of almost \textit{every} object, video captioning usually considers a sequence of frames and biases towards \textit{focused} objects, e.g., the objects that stay in focus regardless of the changing background. Therefore, detecting and properly accommodating focused objects is critical in video captioning. To enforce the description of focused objects and achieve controllable video captioning, we propose an Object-Oriented Non-Autoregressive approach (O2NA), which performs caption generation in three steps: 1) identify the focused objects and predict their locations in the target caption; 2) generate the related attribute words and relation words of these focused objects to form a draft caption; and 3) combine video information to refine the draft caption to a fluent final caption. Since the focused objects are generated and located ahead of other words, it is difficult to apply the word-by-word autoregressive generation process; instead, we adopt a non-autoregressive approach. The experiments on two benchmark datasets, i.e., MSR-VTT and MSVD, demonstrate the effectiveness of O2NA, which achieves results competitive with the state-of-the-arts but with both higher diversity and higher inference speed.

\end{abstract}

\begin{figure}[t]
\centering
\includegraphics[width=1\linewidth]{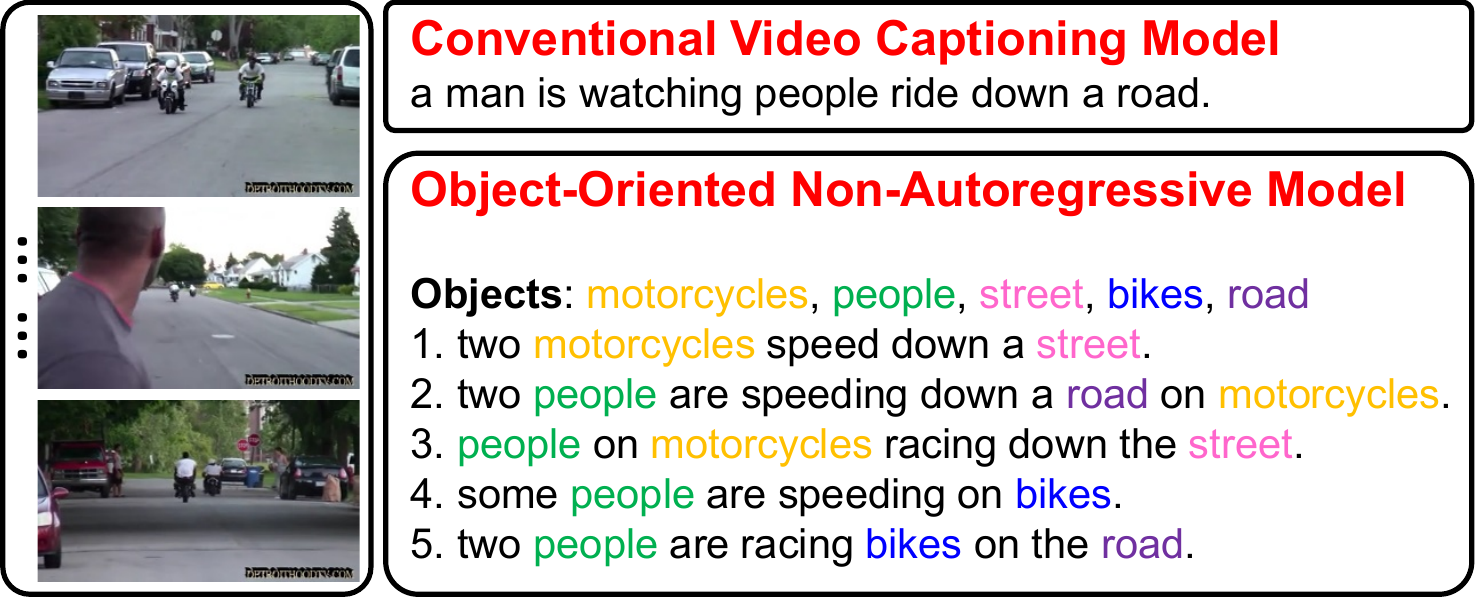}

\caption{Examples of the captions generated by a state-of-the-art conventional video captioning model \cite{Zheng2020SAAT} and our model. Compared to the   conventional model, whose generation process is hardly controllable, our model can be guided to mention the desired objects (i.e., the colored objects) and generate diverse, object-oriented captions for a video.}
\label{fig:compare}
\end{figure}

\section{Introduction}
\label{sec:introduction}
The task of video captioning, which aims to generate a descriptive sentence based on the input video, has a wide range of applications.
In recent years, deep neural models, particularly the models based on the encoder-decoder framework
\cite{Venugopalan2015vc3,Pan2016vc1,Xu2017MA-LSTM,Aafaq2019GRU-EVE}, have achieved great success in advancing the state-of-the-art \cite{Pan2020Spatio,Zheng2020SAAT,Martin2021Improving,yang2021NACF}. These models usually entail the autoregressive property, i.e., conditioning each word on the previously generated words.

In video captioning, one critical step is to detect and include focused objects. As exemplified in Figure~\ref{fig:compare}, when a dangerous situation occurs, a captioning-based blind-aid system should focus on the dangerous objects on the road to alert the visually-impaired people, rather than over-describe the presence of pedestrians or shops nearby. It means that in the above example, \textit{speeding vehicles} should be considered as focused objects and should be mentioned in the generated caption. While people could identify focused objects in video easily \cite{shinn2008object,corbetta2002control,posner1990attention}, existing captioning systems can hardly be controlled to generate focused objects because of their word-to-word generation practice.
Motivated by those observations, we introduce the problem of controllable video captioning in the sense of controlling contents. 

As shown in Figure~\ref{fig:model}, to solve the controllable video captioning problem, we propose the Object-Oriented Non-Autoregressive approach (O2NA).
Different from conventional models that adopt a left-to-right or word-by-word decoding process, O2NA applies a non-autoregressive manner to control the caption generation.
O2NA first detects all objects that appear in the video and then selects the focused objects for the final caption. For example, in the aforementioned blind-aid system, the system would select the dangerous objects \textit{speeding vehicles} in case of an emergency. 
Next, the caption generation process consists of three main steps:  1) locate all focused objects in the proper locations of the target caption; 2) generate the related attribute words and relation words to form a draft caption; and 3) adopt the iterative refinement approach \cite{Ghazvininejad2019Constant-Time,Lee2018Iterative} to proofread and improve the draft caption. 

For each step, as there is no dependency among generated words, the words can be generated in parallel, indicating a fixed computing time regardless of caption length, while computing time of the conventional autoregressive approach is linear with the caption length. For long captions, conventional methods embody high inference latency, which limits their adoption in real-time applications, e.g., blind-aid system \cite{Voykinska2016helpsee} and human-robot interaction \cite{das2017visual}. 
According to our experiments and analyses on two benchmark datasets, i.e., MSR-VTT \cite{Xu2016MSR-VTT} and MSVD (a.k.a. Youtube2Text) \cite{Guadarrama2013MSVD}, our O2NA is able to produce a descriptive and fluent caption which outperforms several existing methods in terms of both accuracy and efficiency.

Overall, the main contributions of this paper are:
\begin{itemize}
    \item We introduce the problem of controllable video captioning in the sense of controlled contents, which has more practical values than the existing studies on syntactic variations.
    
    \item Specifically, we propose the Object-Oriented Non-Autoregressive approach (O2NA) to tackle the controllable video captioning problem by injecting strong control signals conditioned on selected objects, with the benefits of fast and fixed inference time, which are critical for real-time applications.

    \item We evaluate our approach on two datasets. In particular, our O2NA achieves competitive results with the state-of-the-art methods with higher diversity and higher inference speed.

\end{itemize}

The rest of this paper is organized as follows: Section~\ref{sec:related_work} reviews the related work; Section~\ref{sec:approach} introduces the proposed Object-Oriented Non-Autoregressive approach (O2NA) in detail; Section~\ref{sec:experiments} and Section~\ref{sec:analysis} present the experimental results and analyses, respectively; and finally, Section~\ref{sec:conclusions} concludes the paper.

\section{Related Work}
\label{sec:related_work}
In this section, we describe the related work from 1) Video Captioning, 2) Controllable Image Captioning and 3) Non-Autoregressive Decoding.

\subsection{Video Captioning}
\label{sec:related_work_vc}
Recently, a large number of encoder-decoder based neural models have been proposed for video captioning \cite{Venugopalan2015vc3,Yao2015vc4,Pan2016vc1,Pan2016Hierarchical,Xu2017MA-LSTM,Aafaq2019GRU-EVE,Aafaq2020Survey,Zheng2020SAAT,yang2021NACF,Martin2021Improving}.
These methods mainly introduce a convolutional neural network (CNN) \cite{Krizhevsky2012CNN} to encode the video and employ a LSTM \cite{hochreiter1997long} or a Transformer \cite{Zhou2018Transformer} to generate the coherent captions with the attention mechanism \cite{bahdanau2014neural,Pan2016vc1}.
However, these methods lack controllability, i.e., their behaviors can hardly be influenced.
Our model allows an easy way to control the contents of video captions rather than merely syntactic variations in existing studies.

\subsection{Controllable Image Captioning}
\label{sec:related_work_cic}

Different from image captioning \cite{xu2015show,vinyals2015show,lu2017knowing,anderson2018bottom,fenglin2018simnet} that processes a static image with details of almost every appeared object, video captioning considers a sequence of frames which biases towards focused objects.
It is still worth noting that the controllable image captioning has been explored most recently \cite{Cornia2019Control,Chen2020Say,Zheng2019Intention}.
However, all of them are based on autoregressive decoding, i.e., conditioning each word on the previously generated outputs. 
Therefore, to control the generation of image captions, a major challenge 
is to decide the timing to attend to the region-of-interest (i.e., the object we care about).
\citet{Zheng2019Intention} first fixes the cared object and generates the rest captions to its left and right which can only apply to the case with a single cared object. To scale to multiple cared objects, \citet{Cornia2019Control} implements a region pointer mechanism to predict, at each timestep, whether this pointer should be incremented or not;
\citet{Chen2020Say} introduces the abstract scene graph, to control the generation of captions, they proposed graph-based attention and graph updating mechanisms to adaptively select relevant nodes, which contain the concerned objects to generate next word.

In this work, we focus on controllable video captioning, which is a more challenging problem than controllable image captioning.
It is hard for controllable video captioning to construct the same region-of-interests (RoIs) as in \citet{Cornia2019Control} and scene graphs as in \citet{Chen2020Say}.
To this end, based on the non-autoregressive decoding methods in neural machine translation \cite{Gu2018NANMT,Lee2018Iterative,Ghazvininejad2019Constant-Time,Wang2019DD,Shao2019Retrieving}, we propose Object-Oriented  Non-Autoregressive model, which does not need the RoIs in \citet{Cornia2019Control} or scene graphs in \citet{Chen2020Say} to generate controllable video captions.
Moreover, our approach can generate all the objects we care about in parallel, leading to fast generation speed.

It is worth noting that \citet{Wang2019Controllable,Yuan2020Controllable} also introduced the controllable video captioning. However, they devoted to employing Part-of-Speech (POS) information to guide caption generation, which mainly focuses on improving diversity and adjusting the syntactic structure of the captions, instead of constraining the model to generate captions containing the focused objects.

\vspace{-1pt}
\subsection{Non-Autoregressive Decoding}

Most recently, non-autoregressive decoding has received growing attention in the community of neural machine translation (NMT) \cite{Gu2018NANMT,Ghazvininejad2019Constant-Time,Lee2018Iterative,guo2019non,Shao2019Retrieving,Ghazvininejad2020Aligned,Kasai2020Disentangled,Ren2020Study,Haviv2021Can,Hao2021Shared}. 
Such models remove the sequential dependency and can generate all words of a sequence in one step, resulting in high inference efficiency.
Inspired by the success of non-autoregressive decoding, we propose the Object-Oriented Non-Autoregressive model.
As for the network structure, these current non-autoregressive models usually employ a completely empty sequence as the input of decoder to generate the whole sentence in the early stages, which gives a high risk of producing translation errors.
Different from these works, we consider exploiting the objects in the video and propose to first generate an object-oriented coarse-grained caption, and then refine each object word with rich contextual information to generate the whole caption to alleviate the description ambiguity problem.

\section{Approach}
\label{sec:approach}
We first briefly introduce the backgrounds of our approach and then describe the approach in detail.

\subsection{Backgrounds}
\label{sec:backgrounds}
The backgrounds are introduced from the used Video Representations and Basic Module.  

\paragraph{Video Representations}
For video captioning, image and motion features have been widely used. 
Image features are good at illustrating the shapes, the colors and the relationships of the items in the image; Motion features are important for capturing the actions and temporal interactions.
Following \citet{Pei2019MARN}, given a video, $N = 8$ key frames are uniformly sampled to extract image features $I$. Considering both the past and the future contexts, we take each key frame as the center to generate corresponding motion features $M$.
Specifically, for the image features,
we adopt the ResNet-101 \cite{he2016deep} pre-trained on ImageNet \cite{Deng2009ImageNet} to extract the 2048-D image features ${I} \in \R^{N \times d_i}$ ($d_i = 2048$), which are the output of the last convolutional layer.
The motion features are usually given by the 3D CNN \cite{Tran2015C3D}, we adopt the ResNeXt-101 \cite{Hara2018ResNeXt} pre-trained on the Kinetics dataset \cite{Kay2017Kinetics} to extract the 2048-D motion features ${M} \in \R^{N \times d_m}$ ($d_m = 2048$).
In this paper, both features are projected to $d_h = 512$. 
Then, we use the concatenation of the two projected features as the video representations ${V} \in \R^{2N \times d_h}$ to our model.

\begin{figure*}[t]
\centering
\includegraphics[width=0.9\linewidth]{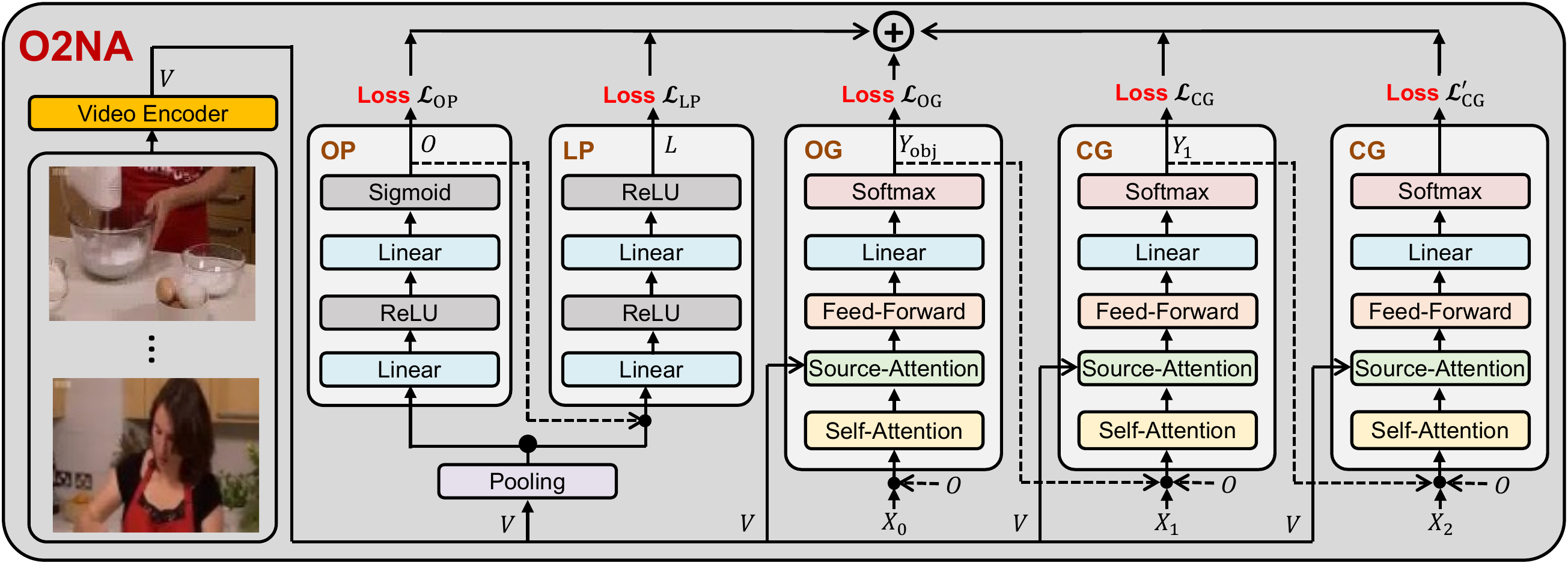}

\caption{Illustration of our proposed O2NA, which consists of an object predictor (OP), a length predictor (LP), an object generator (OG) and a caption generator (CG). The object predictor and length predictor extract the objects appearing to the input video and estimate the length of target caption, respectively; The object generator locates all the focused objects we care about in the target caption;
The caption generator generates the rest words to link focused objects to form a fluent caption.
It is worth noting that the focused objects could be the objects predicted by the object predictor, the preferred objects given by the user or the pre-defined concerned objects, e.g., the dangerous objects in the captioning-based blind-aid system. }
\label{fig:model}
\end{figure*}

\paragraph{Basic Module}
Our approach is adapted from the non-autoregressive decoding models \cite{Lee2018Iterative,Ghazvininejad2019Constant-Time}, which is based on the Transformer decoder (TFM) \cite{ashish2017attention}. Specifically, the TFM consists of a self-attention, a source-attention and a feed-forward network (FF). The multi-head attention (MHA) is the basic of self-attention and source-attention.
Overall, the TFM is defined as follows:
\begin{equation}
\footnotesize
\text{TFM}({Q}, {K}, {V}) = \text{FF}(\text{MHA}(\text{MHA}({Q}, {Q}, {Q}), {K}, {V})) .
\end{equation}
Please refer to \citet{ashish2017attention} for the detailed introduction
of the Transformer decoder (TFM).

\subsection{Object-Oriented Non-Autoregressive Approach (O2NA)}
As stated above, we adopt the Transformer decoder \cite{ashish2017attention} to implement our Object-Oriented Non-Autoregressive approach (O2NA). Specifically, as shown in Figure~\ref{fig:model}, O2NA consists of an object predictor, a length predictor and two Transformer decoders, where the first decoder focuses on generating all the objects we care about in parallel (i.e., object generator), and the second decoder pays attention to linking these objects to form a fluent caption (i.e., caption generator).

\paragraph{Object Predictor (OP)}
The OP is expected to predict the objects that appear in the given video. 
We first build an object vocabulary based on the training captions.
Given this object vocabulary, we can associate each video with a set of objects according to its human-annotated captions. Specifically, we denote the ground truth objects as ${O}^* = \{{o}^*_1, {o}^*_2, \ldots, {o}^*_M\}$, where $M$ represents the size of object vocabulary; ${o}^*_i = 1$ if the video is annotated with object $i$, and ${o}^*_i = 0$ otherwise.
During the training phase, we directly use the ground truth objects ${O}^*$.
At the inference stage, we adopt a two-layer non-linear layer to predict the objects ${O} \in \R^{M}$, defined as:
\begin{equation}
\footnotesize
\begin{aligned}
{O} &= \text{Object-Predictor}(V) \\ 
&= \sigma\left(\text{ReLU}\left(\text{MP}\left({{V}}\right){W}_{\text{O}_1}\right){W}_{\text{O}_2}\right) \\
&\ \text{where} \ \ \text{MP}\left({{V}}\right) = \frac{1}{2N}\sum\nolimits_{i=1}^{2N} v_i ,
\end{aligned}
\end{equation}
where MP denotes the Mean Pooling, $\sigma$ is the sigmoid function; ${W}_{\text{O}_1} \in \R^{d_h \times d_h}$ and ${W}_{\text{O}_2} \in \R^{d_h \times M}$ are the parameters to be learned.
Next, following \citet{wu2016what}, we minimize the element-wise logistic loss function $\mathcal{L}_{\text{OP}}$ to train our OP:
\begin{equation}
\small
\label{eq:obj}
\mathcal{L}_\text{OP}= \sum\nolimits_{i=1}^{M} \log \left(1+\exp \left(-o^*_{i} o_{i}\right)\right)  .
\end{equation}

During the inference procedure, to select the final predicted objects, we set a threshold $\gamma$, which means that if the ${{o}_i} > \gamma$, we reset ${{o}_i} = 1$, and reset ${{o}_i} = 0$ otherwise.
In particular, if we care about some specific objects, for example, the user preferred objects or the pre-defined dangerous objects in the captioning-based blind-aid system, we could just set the value of these concerned objects equal to 1, and set the value of other objects equal to 0.

\paragraph{Length Predictor (LP)}
In the generation process, the non-autoregressive decoding model needs to know the length of target captions \cite{Ghazvininejad2019Constant-Time}.
To this end, at training time, we use the sequence length ${l}^*$ of ground truth caption.
At inference stage, given the video information $V \in \R^{2N \times d_h}$ and the focused objects $O \in \R^{M}$, we adopt a LP to predict the length $l$. 
In detail, we apply a two-layer network to achieve the effect:
\begin{equation}
\footnotesize
\begin{aligned}
l \sim {p}_l &= \text{Length-Predictor}(V, O) \\  &=\text{softmax}\left(\text{ReLU}\left(\left[\text{MP}({{V}}){W}_{\text{L}_V};O W_{\text{L}_O}\right]\right){W}_{\text{L}}\right)  ,
\end{aligned}
\end{equation}
where $[\cdot;\cdot]$ represents the concatenation operation; ${W}_{\text{L}_V} \in \R^{d_h \times d_h}$, ${W}_{\text{L}_O} \in \R^{M \times d_h}$ and ${W}_\text{L} \in \R^{2d_h \times l_{max}}$ are learnable parameters; ${l}_{max} = 30$ denotes the pre-defined maximum sequence length.
Thus, ${p}_l \in \R^{l_{max}}$ is a probability.
We adopt the cross entropy loss $\mathcal{L}_{\text{LP}}$ to train the LP, which can be defined as follows: 
\begin{equation} \label{eq:len}
\footnotesize
\mathcal{L}_{\text{LP}} = - \text{log}({p}_l(l^* | {V}, {O}^*)) .
\end{equation}

\paragraph{Object Generator (OG)}
The object generator is based on the non-autoregressive decoder and is dedicated to generating all the objects we care about at once. To achieve such effect, we adopt a single-layer Transformer decoder\footnote{Our experiments showed that using a single-layer Transformer decoder can achieve the best performance in major metrics with fastest inference speed (Please refer to Section~\ref{sec:layers}).}, 
followed by a linear layer and a softmax function. 
In implementation, the object generator takes the fully masked sequence ${X}_0 = \left(x_{\text{m}_1}, {x_{\text{m}_2}}, \ldots, x_{\text{m}_L}\right), x_{\text{m}_i} \in \R^{d_h}$ with predicted length $l$ by length predictor as input.
The $x_{\text{m}_i} = w_\text{[MASK]} + e_i$, where $w_\text{[MASK]}$ and $e_i$ denotes the word embedding of [MASK] token and position embedding, respectively.
Then the object information ${O}$ is added to ${X}_0$, i.e., $x'_{\text{m}_i} = x_{\text{m}_i} + {O}W_{O}$, where $W_O \in \R^{M \times d_h}$.
At last, the transformer decoder in the object generator takes the ${X}_0 \oplus {O}W_O$ as input ($\oplus$ denotes the matrix-vector addition), and generates all objects at the position in the final caption, i.e., an object-oriented coarse-grained caption, which can be defined as follows:
\begin{equation}
\small
\begin{aligned}
{Y}_\text{obj} \sim {p}_0 &= \text{Object-Generator}({X}_0, {V}, {O}) \\ & = \text{softmax}(\text{TFM}({X}_0 \oplus {O}W_O, {V}, {V})W_\text{OG})  ,
\end{aligned}
\end{equation}
where ${X}_0 \in \R^{l \times d_h}$, ${V} \in \R^{2N \times d_h}$, ${O} \in \R^{M}$ represent the input sequence, the video representations and the predicted objects, respectively; $W_O \in \R^{M \times d_h}$ and $W_\text{OG} \in \R^{d_h \times |D|}$ are the matrices for linear transformation; $|D|$ is the size of vocabulary $D$.
Each value of ${p}_0 \in \R^{l \times |D|}$ is a probability indicating how likely each word in $D$ should be the current output word.

At training time, for each human-annotated caption, we mask all the non-object words based on the object vocabulary to acquire the ground truth object sequence ${Y}^*_\text{obj} = (\ldots, \text{[MASK]}, \ldots, \text{object}_i, \ldots)$.
Our goal is to minimize the following standard cross entropy loss:
\begin{equation}
\small
\label{eq:OG}
\mathcal{L}_\text{OG} = - \sum\nolimits_{i=1}^{l^*} \text{log}({p_0}({y}^{\text{*}}_{\text{obj}_i} | {X}_0, {V}, {O}^*)) .
\end{equation}

\paragraph{Caption Generator (CG)}
In implementation, the caption generator shares the same structure with object generator.
The main differences between the two generators are the different generating objective and the input sequence.
Specifically, the caption generator takes the object sequence ${X}_1$ as input, where ${X}_1$ equals to ${Y}^*_\text{obj}$ and ${Y}_\text{obj}$ at the training stage and inference stage, respectively, and generates the related attribute words and relation words to form a draft caption, which is defined as:
\begin{equation}
\small
\begin{aligned}
{Y}_1 \sim {p}_1 &= \text{Caption-Generator}({X}_1, {V}, {O})  \\ &= \text{softmax}(\text{TFM}({X}_1 \oplus {O}W'_O, {V}, {V})W_\text{CG}) ,
\end{aligned}
\end{equation}
where ${p}_1 \in \R^{l \times |D|}$. Given the ground truth caption ${Y}^*_\text{cap} = (y^*_{\text{cap}_1}, y^*_{\text{cap}_2}, \ldots, y^*_{\text{cap}_l})$,
we adopt standard cross entropy loss as the loss function to train the CG, which can be defined as follows:
\begin{equation} \label{eq:CG}
\small
\mathcal{L}_\text{CG} = - \sum\nolimits_{i=1}^{l^*} \text{log}({p}_1({y}^{\text{*}}_{\text{cap}_i} | {X}_1, {V}, {O}^*)) .
\end{equation}

Since the non-autoregressive approach removes the sequential dependency, we may have introduced the ``multi-modality problem'' \cite{Gu2018NANMT} (i.e., a word could appear in multiple position to form different captions).
So we further adopt the iterative refinement approach \cite{Lee2018Iterative} to proofread ${Y}_1$. 
In implementation, to acquire the input sequence ${X}_2$, we randomly mask $n = \lfloor l*r \rfloor$ words in ${Y}^*_\text{cap}$ and mask out top $n$ words with the lowest confidence in ${Y}_{1}$ at the training time and inference time, respectively, where $l$ and $r$ represent the caption length and masking ratio, respectively, and the confidence is taken to be the output probability. To obtain the final caption, we employ the following equation, which is defined as:
\begin{equation}
\small
\label{eq:CG2_generation}
{Y}_2 \sim {p}_2 = \text{Caption-Generator}({X}_2, {V}, {O}) .
\end{equation}

Finally, the cross entropy loss is defined similar as Eq.~(\ref{eq:CG}):
\begin{equation}
\small
\label{eq:CG2}
\mathcal{L}'_\text{CG} = - \sum\nolimits_{i=1}^{l^*} \text{log}({p}_2({y}^{\text{*}}_{\text{cap}_i} | {X}_2, {V}, {O}^*)) .
\end{equation}

Overall, by combining the $\mathcal{L}_\text{OP}$ in Eq.~(\ref{eq:obj}), $\mathcal{L}_{\text{LP}}$ in Eq.~(\ref{eq:len}), $\mathcal{L}_\text{OG}$ in Eq.~(\ref{eq:OG}), $\mathcal{L}_\text{CG}$ in Eq.~(\ref{eq:CG}) and $\mathcal{L}'_\text{CG}$ in Eq.~(\ref{eq:CG2}), the full training objective is:
\begin{equation}
\small
\label{eq:full_loss}
\mathcal{L}_\text{full} = \lambda_1\mathcal{L}_{\text{LP}} + \lambda_2\mathcal{L}_\text{OP} + \lambda_3\mathcal{L}_\text{OG} + \lambda_4\mathcal{L}_\text{CG} + \lambda_5\mathcal{L}'_\text{CG} ,
\end{equation}
where $\lambda_1, \lambda_2, \lambda_3, \lambda_4$ and  $\lambda_5$ are the hyperparameters that control the regularization.
For simplicity, we set $\lambda_1 = \lambda_2 = \lambda_3 = \lambda_4 = \lambda_5 = 1$, since we find that our approach can achieve competitive results with the state-of-the-art models in major metrics under this setting (see Section~\ref{sec:automatic}), thus we do not attempt to explore other settings.

Overall, through Eq.~(\ref{eq:full_loss}), we are able to realize our Object-Oriented Non-Autoregressive approach (O2NA).
The trained model is encouraged to describe the focused objects that a user cares about.

\section{Experiments}
\label{sec:experiments}

In this section, we first describe the datasets, metrics and settings used for evaluation, then followed by the experimental results of our approach.

\subsection{Datasets, Metrics and Settings}

\subsubsection{Datasets}
Our results are evaluated on the benchmark Microsoft Video Description (MSR-VTT) \cite{Xu2016MSR-VTT} and  Microsoft Video Description (MSVD) \cite{Guadarrama2013MSVD} datasets.
For MSR-VTT, the dataset contains 10,000 video clips, and each video is paired with 20 annotated sentences.
Following common practice \cite{Pei2019MARN,yang2021NACF,Pan2020Spatio}, we use the official splits to report our results. Thus, there are 6513, 497 and 2990 video clips in the training set, validation set and test set, respectively.
For MSVD, it contains 1,970 video clips and roughly 80,000 English sentences. We follow the split settings in \citet{Pei2019MARN}, resulting in 1,200, 100 and 670 videos for the training set, validation set and test set, respectively.
Following previous works, we replace caption words that occur less than 3 times in the training set with the [UNK] token, plus with a [MASK] token, resulting in a vocabulary of 10,546 words for MSR-VTT and 9,467 words for MSVD.

\begin{table*}[t]
\centering
\scriptsize

\setlength{\tabcolsep}{2.5pt}   
\begin{tabular}{@{}l c c c c c c c c c c c c@{}}
\toprule 

\multirow{2}{*}[-3pt]{Methods} & \multicolumn{4}{c}{Dataset: MSVD \cite{Guadarrama2013MSVD}} & \multicolumn{8}{c}{Dataset: MSR-VTT \cite{Xu2016MSR-VTT}} \\ \cmidrule(lr){2-5} \cmidrule(lr){6-13} 
& BLEU-4 & METEOR & ROUGE-L & CIDEr & BLEU-4 & METEOR & ROUGE-L & CIDEr & Novel &Unique &Vocab &VPS \\
\midrule [\heavyrulewidth]

RecNet \cite{Wang2018RecNet_local} & 52.3 & 34.1 & 69.8 & 80.3 & 39.1 & 26.6 & 59.3 & 42.7 &- &- &- &-\\
PickNet \cite{Chen2018PickNet} & 52.3&  33.3&  69.6 & 76.5 & 41.3 & 27.7&  59.8 & 44.1  &- &- &- &-\\
OA-BTG \cite{Zhang2019OA-BTG} & \color{blue} 56.9 & 36.2 & - & 90.6 & 41.4 & 28.2 & - & 46.9  &- &- &- &-\\ 
MARN \cite{Pei2019MARN} & 48.6 & 35.1&  71.9 & 92.2 & 40.4 & 28.1 & 60.7 & 47.1   &- &- &- &-\\
GRU-EVE \cite{Aafaq2019GRU-EVE} & 47.9 & 35.0 & 71.5 & 78.1&  38.3&  28.4 & 60.7 & 48.1 &- &- &- &-\\
POS-Control \cite{Wang2019Controllable} & 52.5&  34.1 & 71.3&  88.7 & 42.0 & 28.2&  61.6&  48.7  &- &- &- &-\\
STAT \cite{Yan2020STAT} & 52.0 & 33.3 & - & 73.8 & 39.3 & 27.1 & - & 43.8  &- &- &- &-\\
STGN-OAKD \cite{Pan2020Spatio} & 52.2&  36.9&  73.9 & 93.0 & 40.5 & 28.3&  60.9 & 47.1  &- &- &- &- \\
ORG-TRL \cite{Zhang2020ORG} & 54.3 & 36.4 & 73.9&  95.2 & \color{blue} 43.6 & \color{blue} 28.8&  62.1 & 50.9  &- &- &- &-\\
SAAT \cite{Zheng2020SAAT} & 46.5 & 33.5&  69.4 & 81.0&  39.9 & 27.7 & 61.2 & 51.0 & \ \ \color{blue}26.8\ssymbol{2} & \ \ \color{blue}35.7\ssymbol{2} & \ \ \color{blue}3.9\ssymbol{2} & \ \ \color{blue}17.6\ssymbol{2} \\
SGN \cite{Ryu2021SGN} &  52.8 & 35.5 & 72.9 & 94.3 & 40.8&  28.3 & 60.8 & 49.5 &- &- &- &- \\
SemSynAN \cite{Martin2021Improving}  & \color{red} 64.4 & \color{red}41.9&  \color{red}79.5 & \color{red}111.5 & \color{red} 46.4 & \color{red} 30.4 & \color{red} 64.7 & \color{red} 51.9  &- &- &- &- \\
\midrule 

O2NA (Ours) & 55.4 & \color{blue} 37.4 & \color{blue} 74.5 & \color{blue}  96.4 & 41.6 & 28.5 & \color{blue} 62.4 & \color{blue} 51.1 & \color{red} 37.2 & \color{red}46.7 & \color{red}4.6 & \color{red}70.8 \\ 
\bottomrule
\end{tabular}
\caption{Performance of automatic evaluation on the test sets of MSVD and MSR-VTT. Higher is better in all columns. \ssymbol{2} denotes our own implementation. VPS stands for videos per second at the inference stage, which is measured on a single NVIDIA GeForce GTX 1080 Ti. 
In this paper, the {\color{red} Red}- and the {\color{blue} Blue}- colored numbers denote the best and the second best results across all approaches, respectively. 
All existing video captioning systems follow the autoregressive approach to generate the captions and cannot control the video captioning process to ensure the inclusion of the focused objects.
In comparison, O2NA can not only describe the focused objects, but also achieve competitive performances with the state-of-the-arts in major metrics with both higher diversity and faster inference.}
\label{tab:result}
\end{table*}

\subsubsection{Metrics}
We test the model performance with a standard captioning evaluation toolkit \cite{chen2015microsoft}. It reports the widely-used automatic evaluation metrics CIDEr \cite{vedantam2015cider}, ROUGE-L \cite{lin2004rouge}, METEOR \cite{lin2003automatic,banerjee2005meteor} and BLEU \cite{papineni2002bleu}. 
Among them, CIDEr, which incorporates the consensus of a reference set for an example, is based on n-gram matching, is specifically designed for evaluating captioning systems.
BLEU and METEOR are originally designed for machine translation evaluation, while ROUGE-L is proposed for automatic evaluation of the extracted text summarization.
Besides, we further adopt the evaluation metrics Novel, Unique and Vocab Usage, provided by \citet{Dai2018Diversity}, to evaluate the diversity of the generated captions. 
Novel is calculated by the percentage of generated captions that have not been seen in the training data; 
Unique is calculated by the percentage of generated unique words among the other all generated captions;
Vocab Usage denotes the percentage of words that are used to generate captions in the vocabulary.

\subsubsection{Settings}
\label{sec:settings}
As stated in Section~\ref{sec:backgrounds}, we set $N=8$, $d_i=d_m=2048$ and $d_h=512$ for the video representations.
All category tags \cite{Xu2016MSR-VTT} included in MSR-VTT.
For the object predictor, to compare with existing methods, we set the threshold $\gamma = 0.8$ and directly select all the predicted objects to generate captions.
For the length predictor, the maximum sequence length $l_{max}$ is set to 30.
For the object generator and caption generator, following the original setting as in Transformer \cite{ashish2017attention}, the model size $d_h =512$. The number of heads in multi-head attention is set to 8 and the feed-forward network dimension is set to 2048. The masking ratio $r = 0.5$.
To build the object vocabulary, we use the spaCy library\footnote{\url{https://spacy.io/}} for noun tagging from the training dataset, resulting in 5,647 and 4,681 noun words for MSR-VTT and MSVD, respectively. The tagged noun words are taken as the object words, building up the object vocabulary with sizes of 5,647 and 4,681 for MSR-VTT and MSVD, respectively. Therefore, we do not use external data to build the object vocabulary.
Specifically, the object predictor labels will match the words used to name objects in the captions.
We use Adam optimizer \cite{kingma2014adam} with a batch size of 64 and a learning rate of 5e-4 within maximum 50 epochs for parameter optimization.

As each video is annotated with multiple sentences, i.e., Video -- \{$\text{Caption}_i$\}, where each sentence $\text{Caption}_i$ includes a set of objects $\{\text{Object}_i\}$, we use all objects appearing in these sentences as the ground truth objects for each video to train the object predictor. However, we treat the different sentences as independent training samples, i.e., Video -- $\text{Caption}_i$ -- $\{\text{Object}_i\}$, to train length predictor, object generator and caption generator.
In this manner, we can ensure that the focused objects $\{\text{Object}_i\}$ appears in the target sentence $\text{Caption}_i$ during training and inference, which allows an easy way to control the contents of video captions.

Following the non-autoregressive decoding models of neural machine translation,
we incorporate the knowledge distillation \cite{Kim2016KD,Gu2018NANMT} and de-duplication \cite{Wang2019DD} techniques to improve the performance of our non-autoregressive model on MSR-VTT.
Furthermore, following \citet{Gu2018NANMT,Wang2019DD,yang2021NACF}, to generate the captions, we also adopt the teacher re-scoring technique and noisy parallel decoding \cite{Gu2018NANMT,yang2021NACF} techniques, which could generate a set of candidate sentences in parallel, then, we select the candidate sentence with the highest output probability as the final generated caption. 
For the detailed introduction of these techniques, please refer to original papers \cite{Kim2016KD,Gu2018NANMT,Wang2019DD,yang2021NACF}.

\subsection{Evaluation Results}
\label{sec:automatic}
In comparable settings, twelve representative methods, including five most recently published state-of-the-art approaches, namely STAT \cite{Yan2020STAT}, STGN-OAKD \cite{Pan2020Spatio}, ORG-TRL \cite{Zhang2020ORG}, SAAT \cite{Zheng2020SAAT}, SGN \cite{Ryu2021SGN} and SemSynAN \cite{Martin2021Improving}, are selected for comparison.
Unless specifically stated, we directly report the results from the original papers.
The results on the test of MSVD and MSR-VTT datasets are shown in Table~\ref{tab:result}.
As we can see, our O2NA achieves the results competitive with the state-of-the-art models on the two datasets in major metrics.
The competitive performances verify the validity of our O2NA for standard video captioning.
More encouragingly, in terms of the metrics that evaluate the diversity of the generated captions, O2NA surpasses the previous state-of-the-art models with relatively 39\%, 31\% and 18\% margins in terms of Novel, Unique and Vocab scores, which proves our arguments and corroborates the effectiveness of our approach.
Moreover, since our O2NA generate the entire captions in three steps with a fixed generation time, we achieve the fastest inference speed (highest VPS in Table~\ref{tab:result}) among existing methods.

Overall, our O2NA achieves performances competitive with state-of-the-arts in major metrics but with higher diversity scores and faster inference speed. The experimental results show that our approach is able to generate fluent and diverse video captions with fast inference speed. 
More importantly, our O2NA allows an easy way to control the contents of video captions rather than merely syntactic variations in existing studies.
These advantages of our approach could have the potential to promote the application of video captioning for real-time industrial applications, e.g., helping visually impaired people see \cite{Voykinska2016helpsee} and human-robot interaction \cite{das2017visual}.

\begin{table*}[t]
\centering
\scriptsize

\setlength{\tabcolsep}{4pt}   
\begin{tabular}{@{}l c c c c c c c c c c c c@{}}
\toprule 

\multirow{2}{*}[-3pt]{Sections} & \multirow{2}{*}[-3pt]{Settings} & \multirow{2}{*}[-3pt]{Methods} & \multirow{2}{*}{\multirow{2}{*}{\begin{tabular}[c]{@{}c@{}} Iteration \\ Times  \end{tabular}}} & \multirow{2}{*}{\multirow{2}{*}{\begin{tabular}[c]{@{}c@{}} Number of \\ Layers  \end{tabular}}} &  \multicolumn{8}{c}{Dataset: MSR-VTT} \\ \cmidrule(lr){6-13} 
&  &  &  & & BLEU-4 & METEOR & ROUGE-L & CIDEr & Novel &Unique &Vocab &VPS \\
\midrule [\heavyrulewidth]

\multirow{3}{*}{\ref{sec:ablation}} & (a) & Baseline & 1 & 1 & 40.0 & 26.9 & 60.2 & 44.6 & 6.6 & 27.1&  2.6 & \color{red}113.7 \\
& (b) & w/ OP & 1 & 1 & 40.7 & 27.4 & 60.6 & 47.9 & 18.0 & 26.9 & 3.1 & 99.5 \\
& \bf O2NA & w/ OP + OG & 1&  1 & 41.6 & 28.5 & 62.4 & 51.1 & \color{red}37.2 & \color{red}46.7 &\color{red} 4.6 & 70.8 \\ \midrule

\multirow{3}{*}{\ref{sec:iteration}} & (c)&  w/ OP + OG & 2 & 1 & 42.1 & 28.7 & 62.5 & 51.6&  31.9&  42.3&  4.0 & 61.0 \\
& (d) & w/ OP + OG & 3 & 1 & 42.4 & \color{red}28.8 & 62.5 & 51.8 & 25.1&  33.0 & 3.5 & 54.9 \\
& (e) & w/ OP + OG & 4 & 1&  \color{red}42.5 & \color{red}28.8 & \color{red}62.6 & \color{red}51.9&  21.1 & 29.3 & 3.0 & 49.3 \\ \midrule

\multirow{3}{*}{\ref{sec:layers}} & (f) & w/ OP + OG & 1 & 2 & 41.8 & 28.5 & 62.1 & 50.8 & 36.0 & 43.7 & 4.5 & 48.5 \\
& (g)&  w/ OP + OG & 1&  3&  41.1&  28.4&  61.5&  50.3&  30.4&  38.6 & 3.9 & 36.9 \\
& (h) & w/ OP + OG & 1 & 4 & 40.5 & 27.6&  61.0 & 48.7 & 22.3&  30.6 & 3.4 & 30.2 \\ 

\bottomrule
\end{tabular}
\caption{Quantitative analysis of O2NA. Baseline denotes the conventional non-autoregressive decoding model in neural machine translation \cite{Lee2018Iterative,Gu2018NANMT,Ghazvininejad2019Constant-Time}. OP and OG denote the object predictor and object generator, respectively.}
\label{tab:ablation}
\end{table*}

\section{Analysis}
\label{sec:analysis}
In this section, we conduct analysis on the benchmark MSR-VTT dataset from different perspectives to better understand our approach.

\subsection{Quantitative Analysis}
\label{sec:quantitative}
We first conduct the quantitative analysis to investigate the contribution of each component in our proposed O2NA.

\subsubsection{Ablation Study}
\label{sec:ablation}
Compared to conventional non-autoregressive decoding models (Baseline) from neural machine translation \cite{Lee2018Iterative,Gu2018NANMT,Ghazvininejad2019Constant-Time}, our O2NA further introduces the \textit{object predictor} and \textit{object generator} for controllable video captioning.
Therefore, we investigate the contribution of the two components and the results are shown in Table~\ref{tab:ablation}.

\paragraph{Effect of the Object Predictor (OP)}
As expected, since the OP can provide explicit visual clues (i.e., objects) of the input video, the model achieves improved results (c.f. Table~\ref{tab:ablation}(b)), especially in Novel and Unique scores, indicating that the OP helps to generate diverse captions. The improved results prove the effectiveness of our OP.

\paragraph{Effect of the Object Generator (OG)}
As shown in Table~\ref{tab:ablation}(O2NA), when further equipping with the OG, the model significantly outperforms the Baseline, which employs a completely empty sequence as the input to generate the whole sentence.
Intuitively, such practice in Baseline may give high risk of producing errors. Fortunately, the object-oriented coarse-grained captions generated by our OG could provide rich contextual information for the following non-autoregressive decoding model to generated accurate revised captions.
It proves our arguments and verifies the effectiveness of generating captions in a coarse-grained to fine-grained manner. 

Overall, the proposed OP and OG can boost the performance from different perspectives, making our O2NA generate diverse and accurate captions. 

\subsubsection{Effect of the Iteration Times}
\label{sec:iteration}
In O2NA, we adopt the iterative refinement technique \cite{Lee2018Iterative} to proofread and improve the generated captions (see Eq.~(\ref{eq:CG2_generation})).
However, in conventional non-autoregressive decoding methods for neural machine translation \cite{Gu2018NANMT,Lee2018Iterative,Ghazvininejad2019Constant-Time,guo2019non,Shao2019Retrieving}, they usually adopt more iterations to obtain better results.
As to O2NA, Table~\ref{tab:ablation}(c-e) shows that performances stabilize with the increasing number of iterations but do not show a significant increase as in  \citet{Lee2018Iterative,Ghazvininejad2019Constant-Time}.
The reason is that our generated object-oriented coarse-grained captions have provided a solid guidance (i.e., rich contextual information) for non-autoregressive video captioning model, which further proves the effectiveness of our approach.
The decreased performance of diversity may be due to the over-fitting problem brought by more iterations, making the model prone to generating frequent captions in the training data. 
Thus, considering the trade-off between ``the performance of caption generation'' and ``the performance of diversity and inference speed'', we only proofread the generated captions once.

\begin{figure*}[t]
\centering
\includegraphics[width=1\linewidth]{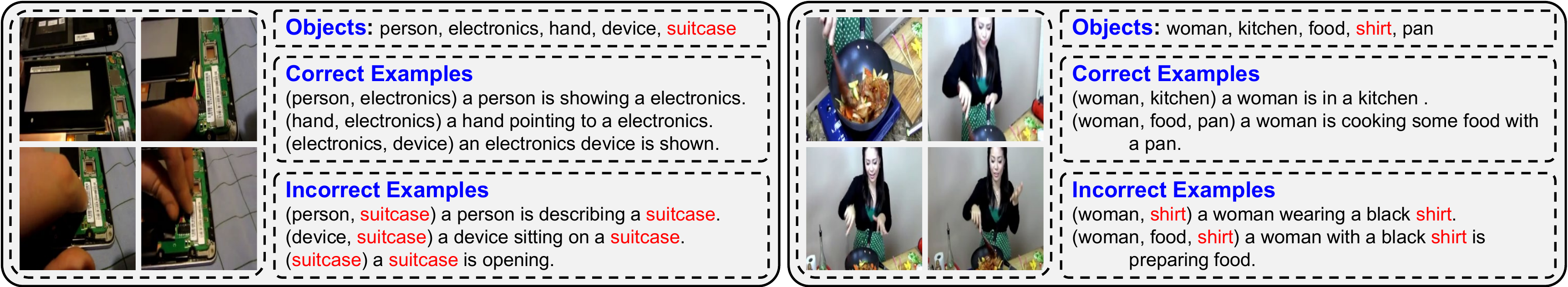}
\caption{Examples of captions generated by our proposed O2NA. For each example, the left plot shows the input video. The upper, middle and lower parts in the right plot show the predicted objects, correct examples and error cases, respectively. The designated objects are listed in brackets. The color Red denotes unfavorable objects.}
\label{fig:example}
\end{figure*}

\subsubsection{Effect of the Number of Layers}
\label{sec:layers}
When increasing the number of layers to 2 (c.f. Table~\ref{tab:ablation}(f)), the model can only achieve a slightly improved result on BLEU-4 (i.e., 41.6 $\to$ 41.8), but loses 31.5\% inference speed.
At the same time, if the number of layers is further increased, the performance decreases. 
We hypothesize that when training on video captioning datasets that are relatively small compared to those for neural machine translation, larger depths add to the difficulty of training, which is the same case with deep RNNs.
In brief, considering the trade-off between the performance and inference speed, we adopt a single-layer Transformer decoder.

\subsection{Case Study and Error Analysis}

In this section, we list some correct and incorrect examples to show the controllability of our proposed O2NA intuitively.
In the analysis, we manually select the predicted objects to encourage the model to generate a set of diverse captions.
Figure~\ref{fig:example} shows that our approach is controllable and explainable. Specifically, it can generate multiple diverse captions for the same video, and can accurately follow the selected objects we care about.
Besides, we find that the error mainly takes place when there are incorrectly predicted objects, e.g.,  ``\textit{suitcase}'' and ``\textit{shirt}''. O2NA mistakes the incorrect object for an appropriate one during its object sequence generation.
A more powerful object predictor may be helpful in solving these problems, but it is unlikely to be completely avoided.

\section{Conclusions}
\label{sec:conclusions}

In this work, we introduce the problem of controllable video captioning in the sense of controlled contents. In contrast to the existing studies considering syntactic variations, controlling contents is of more practical value. 
To tackle the problem, we propose the Object-Oriented Non-Autoregressive approach (O2NA), which encourages the model to describe the focused objects that a user cares about by generating captions conditioned on the focused objects non-autoregressively.
The experiments and analyses verify the flexibility and demonstrate the effectiveness of O2NA, which achieves competitive results with existing state-of-the-art models on two benchmark datasets in major metrics with higher diversity and faster inference.
These advantages could promote the application of video captioning adapting to real-world scenarios.

\section*{Acknowledgments}
This work is supported in part by Beijing Academy of Artificial Intelligence (BAAI).
We sincerely thank all the anonymous reviewers and chairs for their constructive comments and suggestions that substantially improved this paper.
We also sincerely thank Bang Yang for providing the implementation code of non-autoregressive framework for video captioning.\footnote{\url{https://github.com/yangbang18/Non-Autoregressive-Video-Captioning}}
Yuexian Zou and Xu Sun are the corresponding authors of this paper.

\section*{Impact Statement}

This paper introduces the problem of controllable video captioning in the sense of controlled contents to efficiently understand the visual content of a given video and generate corresponding descriptive sentences.
As a result, our work can control the video captioning process and include focused objects, i.e., the video captions generated by our model are more likely to contain preferred objects given by a user or pre-defined objects that should be prioritized in generation.
It improves the practicality of video captioning in real-world applications, such as visual retrieval, human-robot interaction and aiding visually-impaired people.
However, the training of our proposed model relies on a large volume of video-caption pairs, which may not be easily obtained in the real world but could be alleviated using techniques such as distillation from publicly-available pre-trained models.
Hence, it requires specific and appropriate treatment by experienced practitioners.

\bibliographystyle{acl_natbib}
\bibliography{acl2021}

\end{document}